\documentclass[10pt,twocolumn,letterpaper]{article}

\usepackage{cvpr}              

\usepackage{graphicx}
\usepackage{amsmath}
\usepackage{amssymb}
\usepackage{booktabs}

\usepackage[pagebackref,breaklinks,colorlinks]{hyperref}

\usepackage[linesnumbered,ruled,vlined]{algorithm2e}
\usepackage{xcolor}
\usepackage{booktabs}
\usepackage{threeparttable}

\usepackage{amsmath}
\usepackage{multirow}
\usepackage[accsupp]{axessibility}

\DeclareMathOperator*{\argmin}{arg\,min}

\usepackage[capitalize]{cleveref}
\crefname{section}{Sec.}{Secs.}
\Crefname{section}{Section}{Sections}
\Crefname{table}{Table}{Tables}
\crefname{table}{Tab.}{Tabs.}

\def\cvprPaperID{5348} 
\def\confName{CVPR}
\def\confYear{2023}

\begin{document}

\title{Search-Map-Search: A Frame Selection Paradigm for Action Recognition}

\author{Mingjun Zhao\\
University of Alberta\\
Edmonton, Canada\\
{\tt\small zhao2@ualberta.ca}
\and
Yakun Yu\\
Univerity of Alberta\\
Edmonton, Canada\\
{\tt\small yakun2@ualberta.ca }
\and
Xiaoli Wang\\
Tencent\\
Shenzhen, China\\
{\tt\small evexlwang@tencent.com}
\and
Lei Yang\\
Tencent\\
Shenzhen, China\\
{\tt\small yangalalei@gmail.com}
\and
Di Niu\\
Tencent\\
Edmonton, Canada\\
{\tt\small dniu@ualberta.ca}
}

\maketitle

\begin{abstract}
Despite the success of deep learning in video understanding tasks, processing every frame in a video is computationally expensive and often unnecessary in real-time applications. Frame selection aims to extract the most informative and representative frames to help a model better understand video content. 
Existing frame selection methods either individually sample frames based on per-frame importance prediction, without considering interaction among frames,
or adopt reinforcement learning agents to find representative frames in succession, which are costly to train and may lead to potential stability issues.
To overcome the limitations of existing methods, we propose a Search-Map-Search learning paradigm which combines the advantages of heuristic search and supervised learning to select the best combination of frames from a video as one entity. By combining search with learning, the proposed method can better capture frame interactions while incurring a low inference overhead.
Specifically, we first propose a hierarchical search method conducted on each training video to search for the optimal combination of frames with the lowest error on the downstream task. A feature mapping function is then learned to map the frames of a video to the representation of its target optimal frame combination.
During inference, another search is performed on an unseen video to select a combination of frames whose feature representation is close to the projected feature representation. Extensive experiments based on several action recognition benchmarks demonstrate that our frame selection method effectively improves performance of action recognition models, and significantly outperforms a number of competitive baselines. 
\end{abstract}

\section{Introduction}
Videos have proliferated online in recent years with the popularity of social media, and have become a major form of content consumption on the Internet.
The abundant video data has greatly encouraged the development of deep learning techniques for video content understanding. As one of the most important tasks, action recognition aims to identify relevant actions described in videos, and plays a vital role to other downstream tasks like video retrieval and recommendation.

Due to the high computational cost of processing frames in a video, common practices of action recognition involve sampling a subset of frames or clips uniformly \cite{wang2016temporal} or densely \cite{simonyan2014two, yue2015beyond} from a given video a serve as the input to a content understanding model. 
However, since frames in a video may contain redundant information and are not equally important, simple sampling methods are often incapable of capturing such knowledge and hence can lead to sub-optimal action recognition results.

Prior studies attempt to actively select relevant video frames to overcome the limitation of straightforward sampling, achieving improvements to model performance.
Heuristic methods are proposed to rank and select frames according to the importance score of each frame/clip calculated by per-frame prediction \cite{korbar2019scsampler,gowda2021smart}. 
Despite the effectiveness, these methods heavily rely on per-frame features, without considering the interaction or diversity among selected frames.
Reinforcement learning (RL) has also been proposed to identify informative frames by formulating frame selection as a Markov decision process (MDP) \cite{wu2019adaframe,wu2019multi,yeung2016end,fan2018watching,dong2019attention}.
However, existing RL-based methods may suffer from training stability issues and rely on a massive amount of training samples. Moreover, RL methods make an MDP assumption that frames are selected sequentially depending on observations of already selected frames, and thus cannot adjust prior selections based on new observations. 


In this work, we propose a new learning paradigm named Search-Map-Search (SMS), which directly searches for the best combination of frames from a video as one entity.
SMS formulates the problem of frame selection from the perspective of heuristic search in a large space of video frame combinations, which is further coupled with a learnable mapping function to generalize to new videos and achieve efficient inference.

Specifically, we propose a hierarchical search algorithm to efficiently find the most favorable frame combinations on training videos, which are then used as explicit supervision information to train a feature mapping function that maps the feature vectors of an input video to the feature vector of the desirable optimal frame combination.
During inference on an unseen query video, the learned mapping function projects the query video onto a target feature vector for the desired frame combination, where another search process retrieves the actual frame combination that approximates the target feature vector. By combining search with learning, the proposed SMS method can better capture frame interactions while incurring a low inference cost.

The effectiveness of SMS is extensively evaluated on both the long untrimmed action recognition benchmarks, i.e., ActivityNet \cite{caba2015activitynet} and FCVID \cite{jiang2017exploiting}, and the short trimmed UCF101 task \cite{soomro2012ucf101}.
Experimental results show that SMS can significantly improve action recognition models and precisely recognize and produce effective frame selections. Furthermore, SMS significantly outperforms a range of other existing frame selection methods for the same number of frames selected, while can still generate performance higher than existing methods using only 10\% of all labeled video samples for training.
\section{Related Work}

\textbf{Action Recognition}. 
2D ConvNets have been widely utilized for action recognition, where per-frame features are first extracted and later aggregated with different methods such as temporal averaging \cite{wang2016temporal}, recurrent networks \cite{donahue2015long,li2018recurrent,yue2015beyond}, and temporal channel shift \cite{fan2020rubiksnet,lin2019tsm,luo2019grouped}.
Some studies leveraged both the short-term and long-term temporal relationships by two-stream architectures \cite{feichtenhofer2019slowfast,feichtenhofer2016convolutional}.
To jointly capture the spatio-temporal information of videos, 3D ConvNets were proposed, including C3D \cite{tran2015learning}, I3D \cite{carreira2017quo} and X3D \cite{feichtenhofer2020x3d}.
Transformer architecture \cite{vaswani2017attention} have also been applied to video understanding by modeling the spatio-temporal information with attention \cite{bertasius2021space, liu2022video}.

In this paper, we follow the previous frame selection work and apply our method mainly on the temporal averaging 2D ConvNets.

\textbf{Frame Selection.}
The problem of selecting important frames within a video has been investigated in order to improve the performance and reduce the computational cost.

Many researchers focused on selecting frames based on the per-frame heuristic score.
SCSampler \cite{korbar2019scsampler} proposed to select frames based on the predicted scores of a lightweight video model as the usefulness of frames.
SMART \cite{gowda2021smart} incorporated an attention module that takes randomly selected frame pairs as input to model the relationship between frames. 
However, these methods select frames individually regardless of interactions between selected frames, which may lead to redundant selections.

Reinforcement learning (RL) approaches are widely adopted in frame selection to find the effective frames in a trail-and-error setting.
FastForward \cite{fan2018watching} and AdaFrame \cite{wu2019adaframe} adopted a single RL agent to generate a decision on the next frame, and updated the network with policy gradient.
MARL \cite{wu2019multi} formulated the frame sampling process as multiple parallel Markov Decision Processes, and adopted multiple RL agents each responsible for determining a frame.
Although the RL-based approaches are effective, the training stability issue and the requirement of huge amount of training samples with high computational overhead remain a problem.

Recent studies combined frame selection with other techniques to improve the model efficiency.
LiteEval \cite{wu2019liteeval} adopted a two-level feature extraction procedure, where fine expensive features were extracted for important frames, and coarse frames were used for the remaining frames.
ListenToLook \cite{gao2020listen} proposed to use audio information as an efficient video preview for frame selection.
AR-Net \cite{meng2020ar} aimed at selecting the optimal resolution for each frame that is needed to correctly recognize the actions, and learns a differentiable policy using Gumbel Softmax trick \cite{jang2016categorical}.

Our work focuses on the classic task of selecting a subset of frames based on visual information. 
Different from existing methods, our work incorporates a new ``Search-Map-Search'' paradigm that leverages efficient search and supervised feature mapping to explicitly find the best frame combinations, and achieves excellent performance outperforming other frame selection methods.

\section{Methodology}

\begin{figure*}[t]
\centering
\includegraphics[scale=0.78]{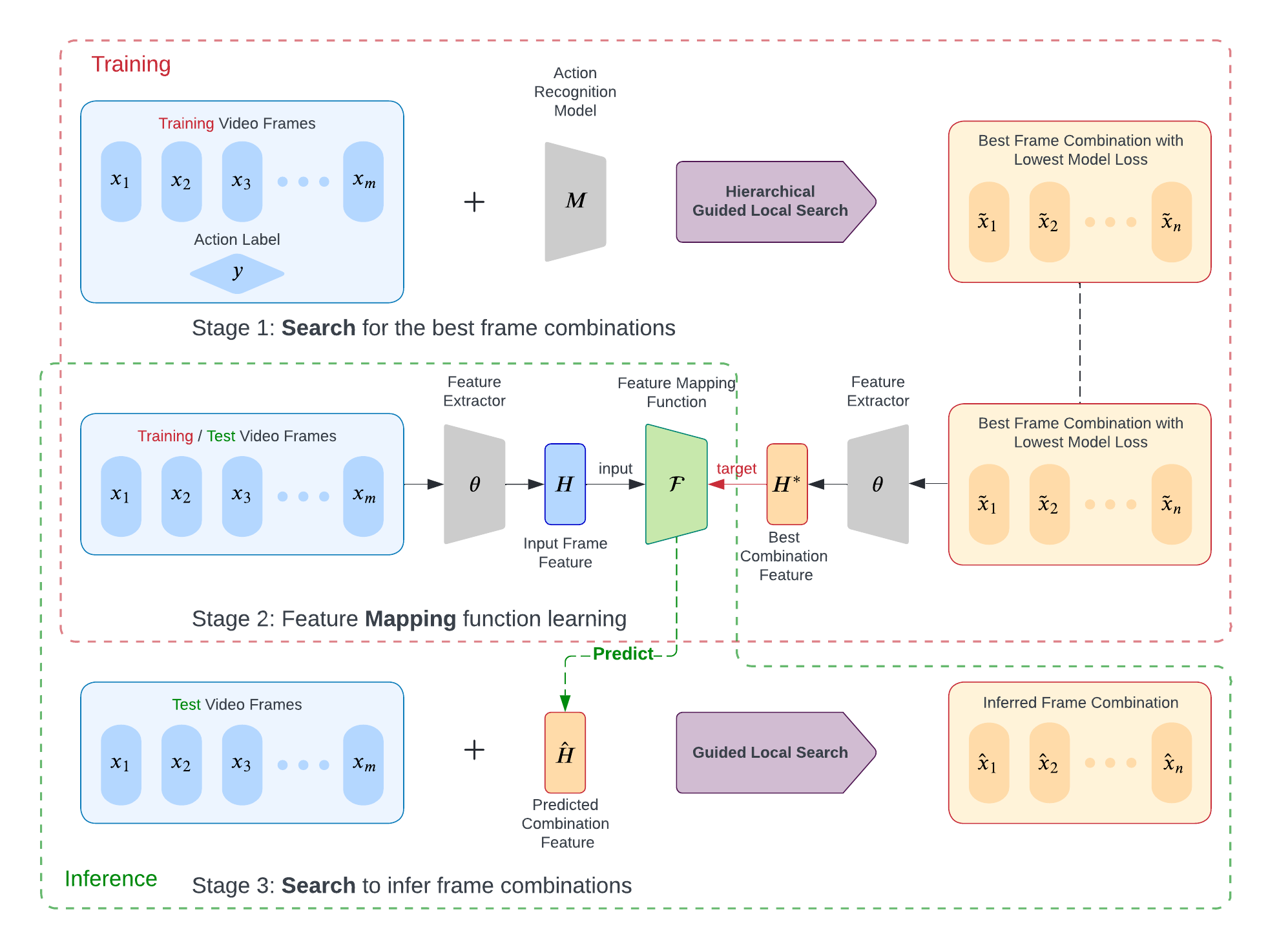}
\caption{An overview of the proposed ``Search-Map-Search'' method. It contains three stages, where in the first stage, an efficient hierarchical search algorithm is used to derive the best frame combinations with lowest losses, which are utilized as the supervised information to train a feature mapping function in the second stage.
In the third stage, for a query video, we incorporate another search process to infer the frame combination whose features are closest to the combination feature predicted with the trained feature mapping function.
}
\label{fig:MethodOverview}
\end{figure*}

\subsection{Overall Architecture}
Figure \ref{fig:MethodOverview} gives an overview of our proposed framework, which consists of three stages: a search stage, a feature mapping stage, and another search stage.
The training process of our method involves the first two stages, while the inference process involves the last two.

Specifically, the goal of the first search stage is to find the best frame combinations in training videos with the lowest model losses, which serve as the supervisory target information for the feature mapping stage. 
We design an efficient hierarchical algorithm coupled with Guided Local Search \cite{voudouris2010guided} to identify the effective frame combinations at a low search cost.

In the second stage, a feature extractor is employed to extract input frame features from the training video frames, and the feature of the best combination from search results.
Then, a feature mapping function is trained via supervised learning by taking the input frame features as input, and transforming it to the target feature of the best combination.

In the third stage, we incorporate another search process to infer the effective frame combination whose feature is closest to the predicted feature from the well-trained feature mapping function.

\subsection{Stage 1: Search for Best Frame Combinations}
Given an action recognition task with a pre-trained model $M$ and a training dataset $D^{tr}=\{X, y\}^{|D^{tr}|}$, where $X=\{x_i\}^{m}_{i=1}$ represents a video sample made up of a collection of $m$ frames, and $y$ is the action label for the video, our goal of stage 1 is to find the best frame combination $\tilde{X}^*$ with $n$ frames for each training video $X$ that minimizes the model loss:
\begin{equation}
    \label{eq:search}
    \begin{aligned}
        \tilde{X}^* &= \argmin_{\tilde{X}} \mathcal{L}\big(M(\tilde{X}), y\big),\\
        \textit{where}\quad \tilde{X} &= \{x_k|x_k \in X\}^n ,
    \end{aligned}
\end{equation}
where $\mathcal{L}$ is the loss function of the action recognition task.
Note that repetitive selection of the same frame is allowed in our setting, as we believe that repeated important frames are better than meaningless frames. 

In order to efficiently find the best frame combinations, we have designed a hierarchical search algorithm 
which exploits the high similarities of adjacent frames by performing search hierarchically on coarse-grained clips first, and then on the fine-grained frames.
Besides, we incorporate Guided Local Search \cite{voudouris2010guided} in our algorithm to exploit per-frame losses as prior information for a good starting search point, which further reduces search costs and empirically outperforms other strong search algorithms such as Genetic Algorithm \cite{whitley1994genetic}.

The overall workflow of our hierarchical search is summarized in Algorithm \ref{alg:hierarchical}.
To begin with, the video is first split into coarse-grained clips each consisting of a collection of non-overlapped frames.
Then, we calculate the model loss for each clip by representing it with the averaged feature vector of all frame inside it, and utilize the information to create an initial solution composed of the clip with the lowest model loss repeated $n$ times.
On top of the initial searching point, we adapt Guided Local Search to find the best clip combination by defining a problem-specific penalty to escape from local optimum points.
The details of Guided Local Search will be introduced in Appendix.
After searching on coarse-grained clips, we again incorporate Guided Local Search to find the best fine-grained frame combinations by replacing each derived clip with a frame inside it.
Via hierarchical search, we have greatly reduced the search space and lowered the search cost significantly, while obtaining satisfying solutions.

\newcommand\mycommfont[1]{\footnotesize\ttfamily\textcolor{blue}{#1}}
\SetCommentSty{mycommfont}

\SetKwInput{KwInput}{Input}
\SetKwInput{KwOutput}{Output}
\SetArgSty{textnormal}

\begin{algorithm}[tb]
\caption{Hierarchical Search}
\label{alg:hierarchical}
\DontPrintSemicolon
\KwInput{Video $X$, Clip Length $K$, Combination Length $n$}
\KwOutput{Frame Combination $\tilde{X}^*$}

\BlankLine \BlankLine
\tcc{clip search phase}
Split the video into clips each consisting of $K$ frames \\
Prepare an initial solution $\tilde{C}_0$ containing the clip with the lowest loss repeated $n$ times \\
Perform Guided Local Search to improve $\tilde{C}_0$ and get $\tilde{C}^*=\{C_k\}_{k=1}^n$ \\

\BlankLine \BlankLine
\tcc{frame search phase}
Define search space for each position in combination $S^F = \{S_k | S_k = \{x_j | x_j \in C_k\}\}_{k=1}^n$ on top of $\tilde{C}^*$ \\
Randomly initialize solution $\tilde{X}_0$ from $S^F$ \\
Perform Guided Local Search to improve $\tilde{X}_0$ and get $\tilde{X}^*=\{x_k\}_{k=1}^n$ \\
\end{algorithm}

\subsection{Stage 2: Feature Mapping Function}
The goal of the second stage is to identify the best frame combination produced in stage 1, given the input video frames by learning a feature mapping function $\mathcal{F}$.

Specifically, the feature mapping function $\mathcal{F}$ takes in the features of input frames $H_0$ generated by a pre-trained feature extractor $\theta$, and outputs a predicted feature $\hat{h} \in \mathcal{R}^d$ representing a frame combination where $d$ denotes the feature dimension:
\begin{equation}
    \label{eq:mapping}
    \begin{aligned}
        \hat{h} &= \mathcal{F}(H_0)\\
        \textit{where}\quad H_0 &= \{h_i | h_i = \theta(x_i)\}_{i=1}^m ,\\
    \end{aligned}
\end{equation}
where $\theta$ is the feature extractor, and $h_i \in \mathcal{R}^d$ is the extracted feature vector for frame $x_i$.

For the network structure of mapping function $\mathcal{F}$, we choose to incorporate transformer layers \cite{vaswani2017attention} to construct the spatio-temporal representations of the input frame features, and an aggregation function to aggregate the representations of variable lengths into a predicted feature vector $\hat{h}$:
\begin{equation}
\begin{aligned}
    H_l &= \textit{transformer}(H_{l-1}) \\
    \hat{h} &= \textit{aggr}(H_l), \\
\end{aligned}
\end{equation}
where $l$ is the number of transformer layers in the mapping function.

The objective of the mapping function is to minimize the distance between the predicted feature $\hat{h}$ and the aggregated feature vector of the searched frame combination $h^*$:
\begin{equation}
    \label{eq:mapping_obj}
    \begin{aligned}
        &\min\quad \textit{dist}(\hat{h}, h^*) \\
        \textit{where}\quad h^* = &\textit{aggr}(\{h_k|h_k=\theta(x_k), x_k \in \tilde{X}^* \}) .\\
    \end{aligned}
\end{equation}


In our implementation, we incorporate cosine distance and mean-pooling as the distance function and aggregation function respectively, while other function choices can be further explored.


\subsection{Stage 3: Search to Infer Frame Combinations}
After the mapping function is learned, it can accurately predict the features of the best frame combinations for unseen videos without relying on the ground truth labels. 
The goal of this stage is to incorporate another search process to infer the frame combinations from the predicted features.
Formally, the objective of the search is to find a frame combination $\hat{X}$ whose aggregated feature $h'$ is closest to the given predicted feature $\hat{h}$:
\begin{equation}
    \label{eq:stage3_obj}
    \begin{aligned}
        \hat{X} &= \argmin_{\tilde{X}} (\textit{dist}(h', \hat{h})) \\
        \textit{where}\quad h' &= \textit{aggr}(\{h_k|h_k=\theta(x_k), x_k \in \tilde{X} \}) .\\
    \end{aligned}
\end{equation}

As the evaluation in the search only involves the calculation of the cosine distance between vectors, which requires little computation, we directly apply Guided Local Search on the fine-grained frame level without the hierarchical setting applied in stage 1.

\section{Experiments}
In this section, we conduct extensive experiments aiming at investigating the following research questions:
\begin{itemize}
    \item \textbf{RQ1:} Can the proposed SMS improve model performance over the base frame sampling method?
    \item \textbf{RQ2:} How does SMS perform compared to other state-of-the-art frame selection methods?
    \item \textbf{RQ3:} What's the computation efficiency of SMS for video inference?
    \item \textbf{RQ4:} How do the different components affect the performance of the proposed SMS?
    \item \textbf{RQ5:} Can SMS generalize well to spatio-temporal models, e.g., transformer based video models?
\end{itemize}

\subsection{Experimental Setup}


\subsubsection{Datasets.}
We evaluate our SMS method on 3 action recognition benchmarks including ActivityNet V1.3 \cite{caba2015activitynet}, FCVID \cite{jiang2017exploiting} and UCF101 \cite{soomro2012ucf101}. 
The videos in ActivityNet and FCVID are untrimmed with average video lengths of $117$ and $167$ seconds respectively, while UCF101 dataset contains trimmed short videos with an average length of $7.21$ seconds.
Table \ref{tab:datset} summarizes the detailed information of the experimental datasets.

\subsubsection{Baselines.}
We compare the proposed SMS with the base selection method and the following state-of-the-art frame selection methods: 
\begin{itemize}
    \item \textbf{Base} is a sparse sampling method proposed in TSN  \cite{wang2016temporal}, where videos are divided into segments of equal length, and frames are randomly sampled within each segments.
    \item \textbf{AdaFrame} \cite{wu2019adaframe} incorporates reinforcement learning to adaptively select informative frames with a memory-augmented LSTM. At testing time, AdaFrame selects different number of frames for each video observed.
    \item \textbf{MARL} \cite{wu2019multi} adopts multiple RL agents each responsible for adjusting the position of a selected frame.
    \item \textbf{SCSampler} \cite{korbar2019scsampler} proposes to select frames based on the prediction scores produced by a lightweight video model.
    \item \textbf{SMART} \cite{gowda2021smart} combines the single-frame predictive score with pair-wise interaction score to make decision on the frame selections.
    \item \textbf{LiteEval} \cite{wu2019liteeval} selects important frames to extract fine features and adopts coarse features for the remaining frames.
    \item \textbf{ListenToLook} \cite{gao2020listen} proposes to use audio information as video preview for frame selection. For a fair comparison, we follow AR-Net \cite{meng2020ar} and include the variant with only the visual modality.
    \item \textbf{AR-Net} \cite{meng2020ar} aims to select the optimal resolutions for frames that are needed to correctly recognize the actions. 
\end{itemize}

\subsubsection{Evaluation metrics.}
Following previous studies, we evaluate the performance of models using mean Average Precision (mAR), which is a commonly adopted metric in action recognition tasks, calculated as the mean value of the average precision over all action classes.

\begin{table}[tb]
\centering
\resizebox{\columnwidth}{!}{
\begin{tabular}{ccccc}
    \toprule
    Dataset & Train & Val & Actions & Avg. Duration \\
    \midrule
    ActivityNet & $10,024$ & $4,926$ & $200$ & $117$s \\
    FCVID & $45,611$ & $45,612$ & $239$ & $167$s \\
    UCF101 & $9,537$ & $3,783$ & $101$ & $7.21$s \\
    \bottomrule
\end{tabular}
}
\caption{Description of evaluation datasets.}
\label{tab:datset}
\end{table}

\begin{table*}[tb]
\centering
\resizebox{\textwidth}{!}{
\begin{tabular}{c|ccc|ccc|cc}
    \toprule
     & \multicolumn{3}{c|}{ActivityNet} & \multicolumn{3}{c|}{FCVID} & \multicolumn{2}{c}{UCF101} \\
    \# Frames & $8$ & $16$ & $25$ & $8$ & $16$ & $25$ & $3$ & $8$ \\
    \midrule
    Base\footnotemark & $77.34\pm0.06$ & $79.41\pm0.05$ & $80.04\pm0.24$ & $83.84\pm0.05$ & $85.34\pm0.03$ & $85.65\pm0.04$ & $90.65\pm0.13$ & $90.70\pm0.10$ \\
    SMS (infer only) & $82.76\pm0.15$ & $83.78\pm0.08$ & $83.85\pm0.17$ & $86.35\pm0.03$ & $86.94\pm0.06$ & $87.08\pm0.04$ & $91.45\pm0.12$ & $91.58\pm0.09$ \\
    SMS (train \& infer) & $\mathbf{83.72\pm0.05}$ & $\mathbf{84.35\pm0.08}$ & $\mathbf{84.56\pm0.08}$ & $\mathbf{86.54\pm0.06}$ & $\mathbf{87.25\pm0.11}$ & $\mathbf{87.59\pm0.01}$ & $\mathbf{91.94\pm0.15}$ & $\mathbf{92.25\pm0.10}$ \\
    \bottomrule
\end{tabular}
}
\caption{Performance comparison of the proposed SMS and the base method. In SMS (infer only), frames are selected with our method only during inference. In SMS (train \& infer), frames are selected with SMS during both training and inference.}
\label{tab:compare_base}
\end{table*}

\begin{table*}[tb]
\centering

\begin{threeparttable}
\begin{tabular}{cc|ccc|ccc}
    \toprule
    \multirow{2}{*}{Method} & \multirow{2}{*}{Backbone} & \multicolumn{3}{c|}{ActivityNet} & \multicolumn{3}{c}{FCVID} \\
     & & \# Frames & mAP & impr. & \# Frames & mAP & impr. \\
    \midrule
    SCSampler & ResNet-50 & $10$ & $72.9$ & $0.4$ & $10$ & $81.0$ & $0.0$ \\
    AdaFrame & ResNet-101 & $8.65$ & $71.5$ & $3.7$ & $8.21$ & $80.2$ & $1.8$ \\
    MARL & ResNet-101 & $8$ & $72.9$ & $0.4$ & - & - & -\\
    SMART & ResNet-101 & $10$ & $73.1$ & - & $10$ & $82.1$ & -\\
    SMART* & ResNet-50 & $8$ & $80.67$ & $3.33$ & $8$ & $83.35$ & $-0.49$\\
    \midrule
    SMS\_10\% & ResNet-50 & $8$ & $82.12$ & $4.78$ & $8$ & $85.69$ & $1.85$\\
    SMS & ResNet-50 & $8$ & $\mathbf{83.72}$ & $\mathbf{6.38}$ & $8$ & $\mathbf{86.54}$ & $\mathbf{2.70}$\\

    \bottomrule
\end{tabular}
\end{threeparttable}
\caption{Performance comparison of the proposed SMS and other state-of-the-art frame selection methods.
We show both their performance and the their reported improvements over the base sampling method, due to the inconsistent base performance among different methods. 
Results of baselines are retrieved from literature, except for SMART*, which is implemented using the same features and implementation settings as SMS. 
We have also included the results of SMS learned only on 10\% training data as SMS\_10\%.
}
\label{tab:compare_other}
\end{table*}

\subsubsection{Implementation details.}
In the first stage of hierarchical searching, the clip length $K$ is set to $30$. 
For the feature mapping network, we adopt a two-layer transformer network with a hidden dimension of $2,048$.
The feature extraction network used in our implementation is a ResNet-50 network \cite{he2016deep} pre-trained on the Kinetics dataset \cite{carreira2017quo}.

For the data pre-processing for action recognition tasks, we decode the video at $1$ fps for long videos in ActivityNet and FCVID and $55$ fps for short videos in UCF101 to retrieve the rgb frames,  which are augmented during training by resizing the short side to $256$, random cropped and resized to $224^2$, after which a random flip with a probability of $0.5$ is applied.
For inference, we resize all frames to $256^2$ and perform three-crop.

For the training of action recognition models, we choose ResNet-50 as the backbone and run $100$ epochs using an SGD optimizer with a momentum of $0.9$, and a step learning rate schedule which decays the learning rate by a factor of $10$ every $40$ epochs.
For ActivityNet and FCVID, we use an initial learning rate of $0.005$, and a batch size of $64$.
For UCF101 with shorter videos, we increase the batch size to $128$ and adjust the initial learning rate to $0.00256$.
All models are trained using code adapted from MMAction2 \cite{2020mmaction2} on eight Nvidia-A100 GPUs.
And we report the average performance and standard deviation of three runs.


\subsection{Effectiveness Analysis (RQ1)}

To validate the effectiveness of the proposed SMS, we make a comprehensive comparison with the base method on both long-video dataset ActivityNet and FCVID, and short-video dataset UCF101.
As the video lengths of different datasets vary in a large range, we choose to select different number of frames for each video in these datasets, where 8, 16 and 25 frames are selected for long videos in ActivityNet and FCVID, and 3 and 8 frames are selected for short videos in UCF101.
Furthermore, we conduct experiments to select frames with SMS only on test data during the inference of pre-trained base models, and incorporate SMS in both the training and inference process.
\footnotetext{The base performance in our paper is higher compared to it in other papers, due to the better codebase and training settings.}
The experimental results are shown in Table \ref{tab:compare_base}, from which we can observe that:
\begin{itemize}
    \item When applied to test data, compared to the base sampling method, SMS (infer only) significantly improves the average mAP on different number of selected frames by $4.53\%$ and $1.85\%$ on ActivityNet and FCVID, respectively. 
    While SMS (infer only) significantly improves model performance, SMS (train \& infer) improves the mAP by $0.75\%$ and $0.34\%$ mAP on ActivityNet and FCVID.
    This observation demonstrates that the frames selected with the SMS method are beneficial to both model training and inference.
    \item While most other frame selection methods only focus on long untrimmed video tasks, SMS is also effective on short trimmed video tasks. Despite that in UCF101, the irrelevant parts of videos are trimmed off, which to a great degree limits the potential of frame selection, SMS still achieves steady improvement of $1.42\%$ average mAP over the base method with the same number of selected frames.
\end{itemize}

\subsection{Performance Comparison (RQ2)}

In order to justify the benefit of our new learning paradigm of frame selection, we compare SMS with other classic frame selection methods on ActivityNet and FCVID.
We choose ResNet as the backbone architecture of the action recognition model and report the performance with similar number of selected frames ranges from $8$ to $10$.

As the implementation and training details are different among the frame selection methods, the performances of the base models are inconsistent.
Therefore, directly comparing the reported performances may be unfair. 
Moreover, due to the lack of source codes and models, we are unable to compare all methods under the same settings, especially for the RL-based methods whose results are difficult to reproduce. 
To make a fairer comparison, in addition to the absolute performance, we also compare the relative improvements of the frame selection methods over the base sampling method.
Also, we have implemented the most recent frame selection method, SMART \cite{gowda2021smart}, using the same features and implementation settings as our proposed SMS, and include its results as SMART*.

\begin{figure}[t]
\centering
\includegraphics[width=\columnwidth]{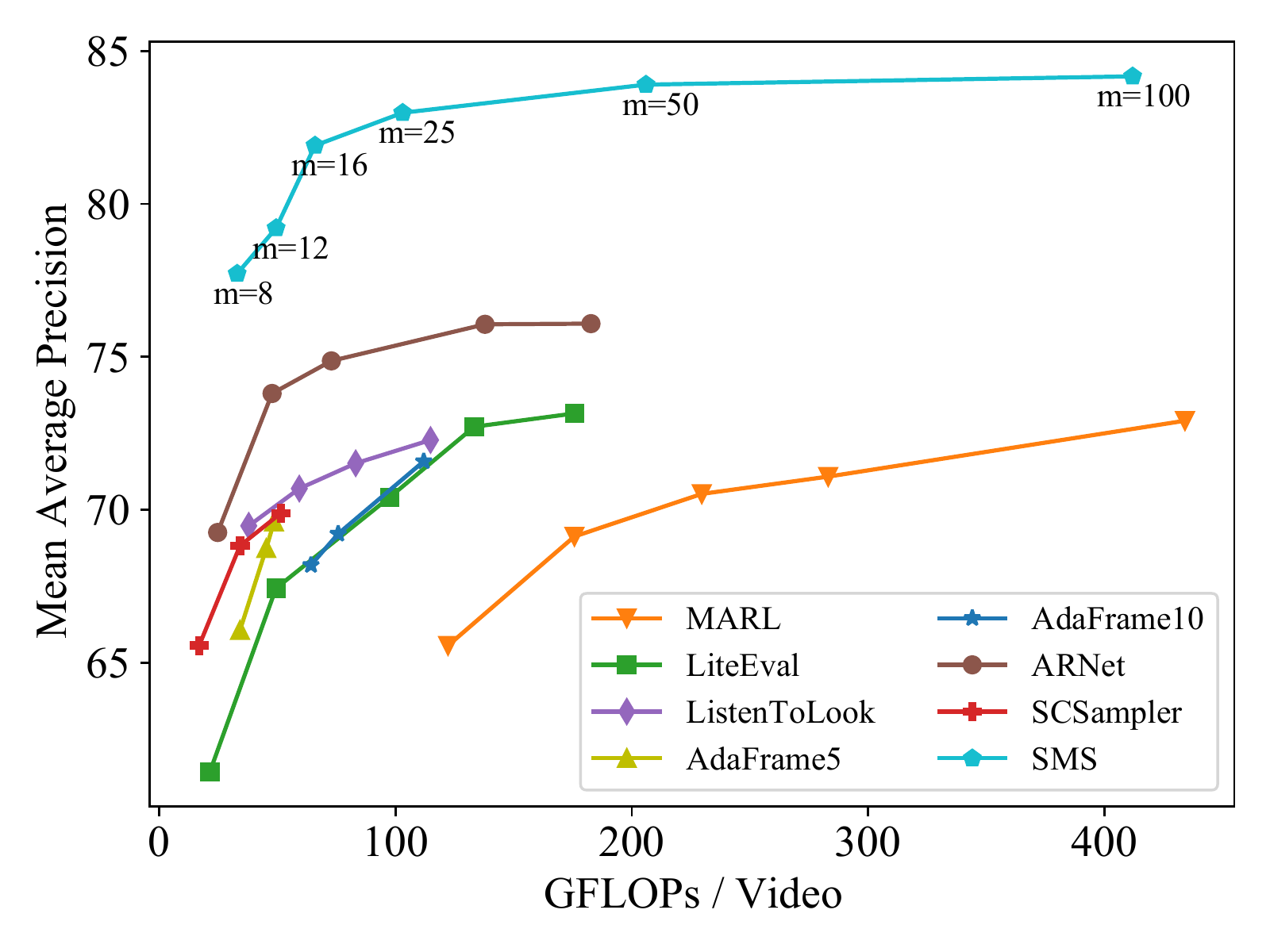}
\caption{
Comparison of SMS with other approaches in terms of performance vs. inference cost (model size per video) evaluated on ActivityNet. We control the inference cost by varying $m$, the number of candidate frames in a video from which $n=8$ best frames are to be selected. 
}
\label{fig:efficiency}
\end{figure}

The comparison results are demonstrated in Table \ref{tab:compare_other}, from which we can see that with the least number of selected frames and the smallest backbone model, the proposed SMS achieves the best performance of $83.72\%$ and $86.54\%$ mAP on ActivityNet and FCVID respectively.
Moreover, even with the higher-performance base models which are harder to improve, 
SMS still obtains the largest improvements of $6.48\%$ and $2.70\%$ among all frame selection methods.
In a fair comparison with the same implementation settings, our method significantly outperforms SMART* by $3.05\%$ and $3.19\%$ respectively on ActivityNet and FCVID.

The reason of our success is due to our novel training paradigm of ``Search-Map-Search'', where an efficient hierarchical search method is incorporated to find the best frame combinations, which better models the frame interactions.
Moreover, in SMS, a feature mapping function is learned to map an input video directly to the optimal frame combination, which is theoretically superior to the one-by-one frame selection adopted by existing methods.

\begin{table*}[tb]
    \centering
    \begin{tabular}{c|c|c|c}
        \toprule
        Feature Extractor (Training Source) & Mapping Model Arch & Performance & Inference GFLOPs \\
        \midrule
        ResNet-50 (Kinetics) & Transformer & $83.72$ & $1.50$ \\
        ResNet-50 (Kinetics) & MLP & $83.22$ & $0.01$ \\
        ResNet-50 (ImageNet) & Transformer & $79.97$ & $1.50$ \\
        ResNet-50 (ActivityNet) & Transformer & $81.06$ & $1.50$ \\
        MobileNet-V2 (Kinetics) & Transformer & $79.53$ & $0.93$ \\
        \bottomrule
    \end{tabular}
    \caption{Ablation analysis results in mAP (\%) on ActivityNet using $8$ frames with different feature extractors and feature mapping model architectures. The feature mapping inference GFLOPs per video is also provided.}
    \label{tab:ablation}
\end{table*}

\begin{table}[tb]
    \centering
    \begin{tabular}{cc|c|c}
        \toprule
        Backbone & \# Frames & Method & Performance \\
        \midrule
        \multirow{4}{*}{TimeSFormer} & \multirow{4}{*}{$8$} & Dense & $84.33$ \\
        & & Base & $90.11$ \\
        & & SMART* & $90.53$ \\
        & & SMS & $\mathbf{91.97}$ \\
        \bottomrule
    \end{tabular}
    \caption{The ActivityNet evaluation results in mAP (\%) using different frame sampling strategies on TimeSFormer.}
    \label{tab:timesformer}
\end{table}




\subsection{Efficiency Analysis (RQ3)}
We now analyze the efficiency of SMS and compare it with other frame selection methods in terms of action recognition performance versus inference cost (evaluated by the model size per video. For a fair comparison, we only include the results that use ResNet as the backbone model.

In order to evaluate the tradeoff between performance and inference efficiency, we first uniformly sub-sample $m$ candidate frames that constitute the search space for each video from which $n=8$ best frames are to be selected. 
As $m$ increases, features are extracted from more candidate frames, and the resulting frame selection becomes more effective, while the inference cost becomes larger.
As demonstrated in \Cref{fig:efficiency}, the action recognition performance of SMS grows rapidly from $m=8$ to $m=25$, while further increasing $m$ to $50$ or $100$ incurs only slight increase in performance but huge inference overhead.
In practice, one can easily achieve a good tradeoff between performance and cost accordingly by tuning the number of candidate frames $m$.

In comparison with other approaches, SMS achieves higher action recognition performance and beats other methods under different computation resource constraints, showing the effectiveness of the proposed search-mapping-search paradigm in frame selection.

\begin{figure}[t]
\centering
\includegraphics[width=0.83\columnwidth]{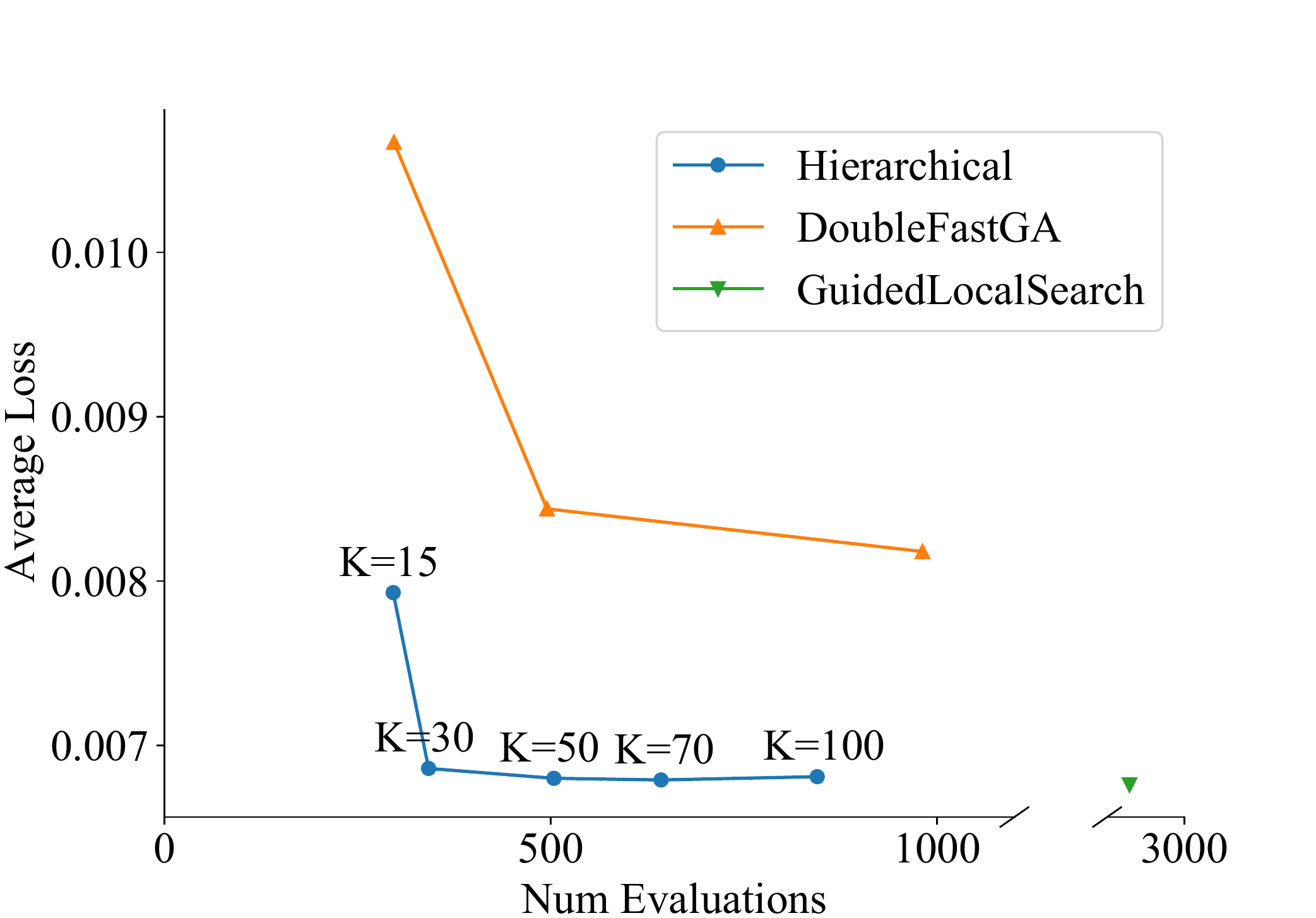}
\caption{
The evaluation of different search algorithms on 100 videos, given by the average loss over the number of evaluations.
}
\label{fig:search_alg}
\end{figure}

\subsection{Ablation Study (RQ4)}
This subsection aims to analyze the effects of different components designed in our proposed SMS.


\textbf{Search algorithm.}
We have conducted experiments to compare the performance and efficiency of our designed Hierarchical Guided Local Search algorithm with other powerful search algorithms such as Fast Genetic Algorithm \cite{doerr2017fast} and the frame-level Guided Local Search \cite{voudouris2010guided}.

In Figure \ref{fig:search_alg}, we show the best loss averaged on $100$ randomly selected videos with different search algorithms, and their corresponding search cost measured by the number of evaluations.
By adopting different clip length $K$, our proposed hierarchical search algorithm can trade off between the search performance and the search cost.
From Figure \ref{fig:search_alg}, we can see that using the same number of evaluations, our hierarchical search achieves better results compared to Fast Genetic Algorithm, by more effectively exploiting the prior knowledge of the per-frame loss information.
The original Guided Local Search without hierarchical design is extremely costly which requires nearly $3,000$ evaluations per video.
Contrastively, our algorithm is more efficient with the hierarchical design, and achieves comparable performance with far less computation cost. 


\textbf{Feature mapping network.}
The process of feature mapping aims to transform the input frame features to the feature of target combination.
We have conducted experiments to explore the impact of different network architectures and training data sources for feature mapping.

For the network architecture, we adopt transformers to sequentially modeling the frame features with their spatio-temporal relations taken into consideration.
Another simple applicable design is to adopt a mean-pooling layer that aggregates all the features of frames into a single feature vector, followed by a simpler two-layer MLP network.
In Table \ref{tab:ablation}, row 1 and 2 compares the performance and the inference efficiency of the two designs. 
As we can see, using transformer model as the mapping function outperforms MLP design by $0.5\%$ mAP due to the better representation ability, while the inference cost of both designs are small and negligible (less than 2 GFLOPs) compared to the cost of frame feature extractions (tens or hundreds GFLOPs).

\textbf{Feature extractor.}
We have analyzed the effect of different feature extractor settings by trying smaller model structure, e.g., MobileNet-V2 \cite{sandler2018mobilenetv2}, and different pre-trained data sources including ImageNet \cite{deng2009imagenet}, Kinetics \cite{carreira2017quo} and ActivityNet \cite{caba2015activitynet}.

Comparing the results of row 1, 3 and 4 in Table \ref{tab:ablation}, we can see that the pre-training data of feature extractor can make a difference.
The feature extractor trained on the largest Kinetics dataset achieves the best performance as it better captures the semantics of actions by training on more related samples, compared to the ones trained on smaller ActivityNet dataset and out-domain ImageNet dataset.
Besides, as shown in row 5 of Table \ref{tab:ablation}, using smaller models such as MobileNet-V2 for extractor can lead to performance decline.
In general, the representation capability of feature extractors is valuable for frame selection to recognize the important frames and find the best frame combinations.

\subsection{Generalizability Analysis (RQ5)}
It is a natural question to ask if the frames selected by SMS can also be beneficial to spatio-temporal video models.
To find out the answer to this question, we have conducted an experiment to apply the SMS selected frames on TimeSFormer \cite{bertasius2021space}, which incorporates the transformer architecture and is one of the most advanced video models.

The results are shown in Table \ref{tab:timesformer}.
The dense frame sampling strategy incorporated in the original TimeSFormer implementation randomly samples a clip containing $8$ successive frames from videos, and achieves $84.33\%$ mAP on ActivityNet.
However, in untrimmed video dataset, dense sampling can only captures a small part of the video and may miss important information.
In contrast, the base sampling strategy selects frames uniformly from videos and achieves $90.11\%$ mAP, while SMART* achieves $90.53\%$ mAP.

By using the input frames selected by SMS, we obtain a significant performance gain of $1.86\%$ over the base sampling strategy and $1.44\%$ over SMART*, and achieve $91.97\%$ mAP.
This improvement demonstrates the strong generalizability of SMS across different model architectures, and that SMS is not only effective on 2D video modeling, but can also capture the spatio-temporal relationship among frames and is beneficial to 3D video model learning.

\section{Conclusion}
In this paper, we propose a new learning paradigm for frame selection, called ``Search-Map-Search'', which consists of a search stage to efficiently find the best frame combination with a hierarchical search algorithm, a feature mapping stage that learns to transform the input frame features directly into the feature of the searched combination, and another search stage that selects frames based on the mapped feature.
Compared with existing frame selection methods, SMS is a more accurate learning paradigm that takes advantage of efficient search and supervised feature mapping to directly select the best combination of frames as one entity, which better captures the frame interactions.
Experimental results show that SMS achieves significant performance gains on multiple action recognition benchmarks, and outperforms other strong baseline methods.

{\small
\bibliographystyle{ieee_fullname}
\bibliography{main}

\begin{thebibliography}{10}\itemsep=-1pt

\bibitem{bertasius2021space}
Gedas Bertasius, Heng Wang, and Lorenzo Torresani.
\newblock Is space-time attention all you need for video understanding?
\newblock In {\em ICML}, volume~2, page~4, 2021.

\bibitem{caba2015activitynet}
Fabian Caba~Heilbron, Victor Escorcia, Bernard Ghanem, and Juan Carlos~Niebles.
\newblock Activitynet: A large-scale video benchmark for human activity
  understanding.
\newblock In {\em Proceedings of the ieee conference on computer vision and
  pattern recognition}, pages 961--970, 2015.

\bibitem{carreira2017quo}
Joao Carreira and Andrew Zisserman.
\newblock Quo vadis, action recognition? a new model and the kinetics dataset.
\newblock In {\em proceedings of the IEEE Conference on Computer Vision and
  Pattern Recognition}, pages 6299--6308, 2017.

\bibitem{2020mmaction2}
MMAction2 Contributors.
\newblock Openmmlab's next generation video understanding toolbox and
  benchmark.
\newblock \url{https://github.com/open-mmlab/mmaction2}, 2020.

\bibitem{deng2009imagenet}
Jia Deng, Wei Dong, Richard Socher, Li-Jia Li, Kai Li, and Li Fei-Fei.
\newblock Imagenet: A large-scale hierarchical image database.
\newblock In {\em 2009 IEEE conference on computer vision and pattern
  recognition}, pages 248--255. Ieee, 2009.

\bibitem{doerr2017fast}
Benjamin Doerr, Huu~Phuoc Le, R{\'e}gis Makhmara, and Ta~Duy Nguyen.
\newblock Fast genetic algorithms.
\newblock In {\em Proceedings of the Genetic and Evolutionary Computation
  Conference}, pages 777--784, 2017.

\bibitem{donahue2015long}
Jeffrey Donahue, Lisa Anne~Hendricks, Sergio Guadarrama, Marcus Rohrbach,
  Subhashini Venugopalan, Kate Saenko, and Trevor Darrell.
\newblock Long-term recurrent convolutional networks for visual recognition and
  description.
\newblock In {\em Proceedings of the IEEE conference on computer vision and
  pattern recognition}, pages 2625--2634, 2015.

\bibitem{dong2019attention}
Wenkai Dong, Zhaoxiang Zhang, and Tieniu Tan.
\newblock Attention-aware sampling via deep reinforcement learning for action
  recognition.
\newblock In {\em Proceedings of the AAAI Conference on Artificial
  Intelligence}, volume~33, pages 8247--8254, 2019.

\bibitem{fan2018watching}
Hehe Fan, Zhongwen Xu, Linchao Zhu, Chenggang Yan, Jianjun Ge, and Yi Yang.
\newblock Watching a small portion could be as good as watching all: Towards
  efficient video classification.
\newblock In {\em IJCAI International Joint Conference on Artificial
  Intelligence}, 2018.

\bibitem{fan2020rubiksnet}
Linxi Fan, Shyamal Buch, Guanzhi Wang, Ryan Cao, Yuke Zhu, Juan~Carlos Niebles,
  and Li Fei-Fei.
\newblock Rubiksnet: Learnable 3d-shift for efficient video action recognition.
\newblock In {\em European Conference on Computer Vision}, pages 505--521.
  Springer, 2020.

\bibitem{feichtenhofer2020x3d}
Christoph Feichtenhofer.
\newblock X3d: Expanding architectures for efficient video recognition.
\newblock In {\em Proceedings of the IEEE/CVF Conference on Computer Vision and
  Pattern Recognition}, pages 203--213, 2020.

\bibitem{feichtenhofer2019slowfast}
Christoph Feichtenhofer, Haoqi Fan, Jitendra Malik, and Kaiming He.
\newblock Slowfast networks for video recognition.
\newblock In {\em Proceedings of the IEEE/CVF international conference on
  computer vision}, pages 6202--6211, 2019.

\bibitem{feichtenhofer2016convolutional}
Christoph Feichtenhofer, Axel Pinz, and Andrew Zisserman.
\newblock Convolutional two-stream network fusion for video action recognition.
\newblock In {\em Proceedings of the IEEE conference on computer vision and
  pattern recognition}, pages 1933--1941, 2016.

\bibitem{gao2020listen}
Ruohan Gao, Tae-Hyun Oh, Kristen Grauman, and Lorenzo Torresani.
\newblock Listen to look: Action recognition by previewing audio.
\newblock In {\em Proceedings of the IEEE/CVF Conference on Computer Vision and
  Pattern Recognition}, pages 10457--10467, 2020.

\bibitem{gowda2021smart}
Shreyank~N Gowda, Marcus Rohrbach, and Laura Sevilla-Lara.
\newblock Smart frame selection for action recognition.
\newblock In {\em Proceedings of the AAAI Conference on Artificial
  Intelligence}, volume~35, pages 1451--1459, 2021.

\bibitem{he2016deep}
Kaiming He, Xiangyu Zhang, Shaoqing Ren, and Jian Sun.
\newblock Deep residual learning for image recognition.
\newblock In {\em Proceedings of the IEEE conference on computer vision and
  pattern recognition}, pages 770--778, 2016.

\bibitem{jang2016categorical}
Eric Jang, Shixiang Gu, and Ben Poole.
\newblock Categorical reparameterization with gumbel-softmax.
\newblock {\em arXiv preprint arXiv:1611.01144}, 2016.

\bibitem{jiang2017exploiting}
Yu-Gang Jiang, Zuxuan Wu, Jun Wang, Xiangyang Xue, and Shih-Fu Chang.
\newblock Exploiting feature and class relationships in video categorization
  with regularized deep neural networks.
\newblock {\em IEEE transactions on pattern analysis and machine intelligence},
  40(2):352--364, 2017.

\bibitem{korbar2019scsampler}
Bruno Korbar, Du Tran, and Lorenzo Torresani.
\newblock Scsampler: Sampling salient clips from video for efficient action
  recognition.
\newblock In {\em Proceedings of the IEEE/CVF International Conference on
  Computer Vision}, pages 6232--6242, 2019.

\bibitem{li2018recurrent}
Dong Li, Zhaofan Qiu, Qi Dai, Ting Yao, and Tao Mei.
\newblock Recurrent tubelet proposal and recognition networks for action
  detection.
\newblock In {\em Proceedings of the European conference on computer vision
  (ECCV)}, pages 303--318, 2018.

\bibitem{lin2019tsm}
Ji Lin, Chuang Gan, and Song Han.
\newblock Tsm: Temporal shift module for efficient video understanding.
\newblock In {\em Proceedings of the IEEE/CVF International Conference on
  Computer Vision}, pages 7083--7093, 2019.

\bibitem{liu2022video}
Ze Liu, Jia Ning, Yue Cao, Yixuan Wei, Zheng Zhang, Stephen Lin, and Han Hu.
\newblock Video swin transformer.
\newblock In {\em Proceedings of the IEEE/CVF Conference on Computer Vision and
  Pattern Recognition}, pages 3202--3211, 2022.

\bibitem{luo2019grouped}
Chenxu Luo and Alan~L Yuille.
\newblock Grouped spatial-temporal aggregation for efficient action
  recognition.
\newblock In {\em Proceedings of the IEEE/CVF International Conference on
  Computer Vision}, pages 5512--5521, 2019.

\bibitem{meng2020ar}
Yue Meng, Chung-Ching Lin, Rameswar Panda, Prasanna Sattigeri, Leonid
  Karlinsky, Aude Oliva, Kate Saenko, and Rogerio Feris.
\newblock Ar-net: Adaptive frame resolution for efficient action recognition.
\newblock In {\em European Conference on Computer Vision}, pages 86--104.
  Springer, 2020.

\bibitem{sandler2018mobilenetv2}
Mark Sandler, Andrew Howard, Menglong Zhu, Andrey Zhmoginov, and Liang-Chieh
  Chen.
\newblock Mobilenetv2: Inverted residuals and linear bottlenecks.
\newblock In {\em Proceedings of the IEEE conference on computer vision and
  pattern recognition}, pages 4510--4520, 2018.

\bibitem{simonyan2014two}
Karen Simonyan and Andrew Zisserman.
\newblock Two-stream convolutional networks for action recognition in videos.
\newblock {\em Advances in neural information processing systems}, 27, 2014.

\bibitem{soomro2012ucf101}
Khurram Soomro, Amir~Roshan Zamir, and Mubarak Shah.
\newblock Ucf101: A dataset of 101 human actions classes from videos in the
  wild.
\newblock {\em arXiv preprint arXiv:1212.0402}, 2012.

\bibitem{tran2015learning}
Du Tran, Lubomir Bourdev, Rob Fergus, Lorenzo Torresani, and Manohar Paluri.
\newblock Learning spatiotemporal features with 3d convolutional networks.
\newblock In {\em Proceedings of the IEEE international conference on computer
  vision}, pages 4489--4497, 2015.

\bibitem{vaswani2017attention}
Ashish Vaswani, Noam Shazeer, Niki Parmar, Jakob Uszkoreit, Llion Jones,
  Aidan~N Gomez, {\L}ukasz Kaiser, and Illia Polosukhin.
\newblock Attention is all you need.
\newblock {\em Advances in neural information processing systems}, 30, 2017.

\bibitem{voudouris2010guided}
Christos Voudouris, Edward~PK Tsang, and Abdullah Alsheddy.
\newblock Guided local search.
\newblock In {\em Handbook of metaheuristics}, pages 321--361. Springer, 2010.

\bibitem{wang2016temporal}
Limin Wang, Yuanjun Xiong, Zhe Wang, Yu Qiao, Dahua Lin, Xiaoou Tang, and
  Luc~Van Gool.
\newblock Temporal segment networks: Towards good practices for deep action
  recognition.
\newblock In {\em European conference on computer vision}, pages 20--36.
  Springer, 2016.

\bibitem{whitley1994genetic}
Darrell Whitley.
\newblock A genetic algorithm tutorial.
\newblock {\em Statistics and computing}, 4(2):65--85, 1994.

\bibitem{wu2019multi}
Wenhao Wu, Dongliang He, Xiao Tan, Shifeng Chen, and Shilei Wen.
\newblock Multi-agent reinforcement learning based frame sampling for effective
  untrimmed video recognition.
\newblock In {\em Proceedings of the IEEE/CVF International Conference on
  Computer Vision}, pages 6222--6231, 2019.

\bibitem{wu2019liteeval}
Zuxuan Wu, Caiming Xiong, Yu-Gang Jiang, and Larry~S Davis.
\newblock Liteeval: A coarse-to-fine framework for resource efficient video
  recognition.
\newblock {\em Advances in Neural Information Processing Systems}, 32, 2019.

\bibitem{wu2019adaframe}
Zuxuan Wu, Caiming Xiong, Chih-Yao Ma, Richard Socher, and Larry~S Davis.
\newblock Adaframe: Adaptive frame selection for fast video recognition.
\newblock In {\em Proceedings of the IEEE/CVF Conference on Computer Vision and
  Pattern Recognition}, pages 1278--1287, 2019.

\bibitem{yeung2016end}
Serena Yeung, Olga Russakovsky, Greg Mori, and Li Fei-Fei.
\newblock End-to-end learning of action detection from frame glimpses in
  videos.
\newblock In {\em Proceedings of the IEEE conference on computer vision and
  pattern recognition}, pages 2678--2687, 2016.

\bibitem{yue2015beyond}
Joe Yue-Hei~Ng, Matthew Hausknecht, Sudheendra Vijayanarasimhan, Oriol Vinyals,
  Rajat Monga, and George Toderici.
\newblock Beyond short snippets: Deep networks for video classification.
\newblock In {\em Proceedings of the IEEE conference on computer vision and
  pattern recognition}, pages 4694--4702, 2015.

\end{thebibliography}
}

\end{document}